\documentclass[letterpaper]{article} 
\usepackage{aaai2027}  
\usepackage[hyphens]{url}  
\usepackage{graphicx} 
\usepackage[numbers,sort&compress]{natbib}
\usepackage{caption}  
\usepackage{algorithm}
\usepackage{algpseudocode}
\usepackage{booktabs}
\usepackage{amsmath}
\usepackage{amssymb}
\usepackage{tabularx}
\usepackage{array}

\graphicspath{{figures/}}

\newcommand{\method}{GradReason}

\newcommand{\pmstd}[1]{\,{\scriptsize$\pm$#1}}
\newcommand{\ci}[2]{\,{\scriptsize[$#1$,\,$#2$]}}
\newcommand{\binv}{b_{\mathrm{inv}}}
\newcommand{\bprog}{b_{\mathrm{prog}}}
\newcolumntype{Y}{>{\raggedright\arraybackslash}X}

\title{\method{}: LLMs as Meta-Learners that Replace Training Epochs with Training-Data-Free Weight Updates}
\author{
    Ali Mohammadi Ruzbahani\textsuperscript{\rm 1},
    Abbas Yazdinejad\textsuperscript{\rm 2},
    Hadis Karimipour\textsuperscript{\rm 1}
}
\affiliations{
    \textsuperscript{\rm 1}Smart Cyber-Physical Systems (SCPS) Lab, University of Calgary, Calgary, Alberta, Canada\\
    \textsuperscript{\rm 2}Decentralized Cybersecurity and Artificial Intelligence(DCAILab), University of Regina, Regina, Saskatchewan, Canada\\
    ali.mohammadiruzbaha@ucalgary.ca, Abbas.Yazdinejad@uregina.ca, hadis.karimipour@ucalgary.ca
}

\begin{document}

\maketitle

\begin{abstract}
Standard training recipes make a fixed number of passes over the training data, even when an epoch mostly adds mini-batch noise. We study whether a large language model (LLM) that reads the weight and gradient history of a run can replace some of these epochs with updates that compute no gradient and read no training image. In \method{}, an LLM agent is called every ten epochs. In a restricted sandbox it analyses recent weight snapshots, epoch-mean gradients and validation metrics, then submits a Python program that computes new weights and a plan marking each of the next ten epochs as data or virtual. A virtual epoch is kept only if it passes a forward-only check on 5,000 held-out validation images. In 120 runs on CIFAR-10 and CIFAR-100 with three architectures, Claude Opus 5.5 used 32.8\% fewer data epochs than the 80-epoch recipe at 0.31 percentage points lower test accuracy (95\% CI $-0.72$ to $+0.09$; 0.94 points on ResNet-18), on the recipe's own frontier of data against accuracy. Unprompted, Opus averaged recent weight snapshots in every run. This choice produced the savings: a fixed rule with the same acceptance check but extrapolations from prior work saved 5.4\% of the data epochs, and the same rule with averaging added saved 25.0\%. The LLM's contribution is to find a working update within a single run; once found, a fixed rule implements it equally well. An Opus run cost 0.68 USD, but at this scale the agent's latency exceeded the GPU time it saved.
\end{abstract}

\section{Introduction}
\label{sec:intro}

This paper asks whether a large language model (LLM) that reads the weight and gradient history of a training run can replace some passes over the training data by updates that use no training data, and still reach the accuracy of the standard recipe \cite{a1}. A standard recipe fixes the optimiser, the learning-rate schedule and the number of epochs before training starts, and it spends every epoch on a full pass over the data. Early in a cosine schedule, at a high learning rate, the mean gradients of consecutive epochs of a ResNet-18 on CIFAR-10 are almost orthogonal (cosine about 0.03 in our runs). An epoch then moves the weights a long way, but mostly through mini-batch noise. A run also produces a history of weight snapshots and gradients that the optimiser discards, apart from its momentum buffers. To use it, a model has to decide during the run when a data epoch can be replaced by a cheaper update and what that update should be. We impose one constraint: a replacement may not compute a gradient on any data. It may only do arithmetic on the recorded weights, gradients and optimiser state, and it is checked by a few forward passes on a small validation set. We call such an update \emph{training-data-free}. It is not free of data altogether: the 5{,}000 held-out validation images are used, in forward passes only, to accept or reject each update and to recalibrate batch normalisation. They are carved out of the training set for every method, including the recipe, so no method trains on more images than another, and we charge their forward passes to the method's GPU time.

\begin{figure*}[t]
  \centering
  \includegraphics[width=\textwidth]{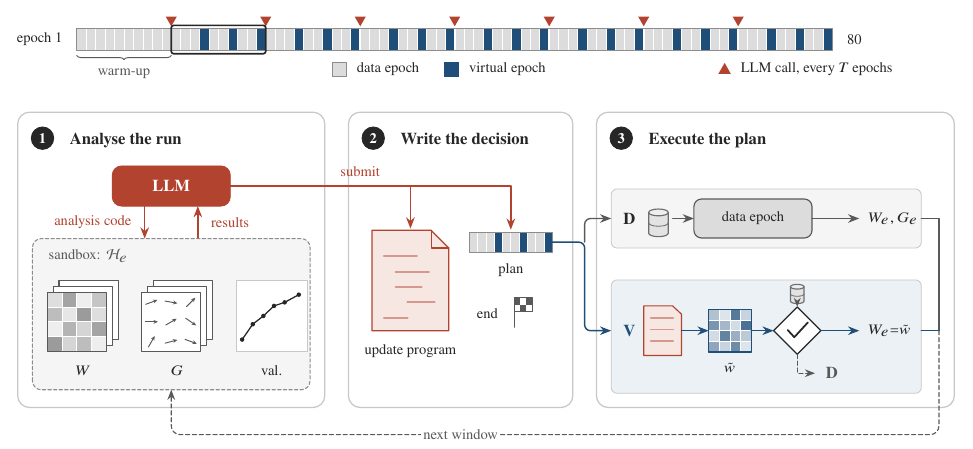}
  \caption{Overview of \method{}: the epoch clock (top) and one window (bottom).}
  \label{fig:overview}
\end{figure*}

Our idea is to make the LLM the meta-learner of the run and to let it express each decision as a short program. In \method{} (Fig.~\ref{fig:overview}), training starts with ten ordinary data epochs. From then on the LLM is called every ten epochs in a new conversation. It examines the last ten weight snapshots, the epoch-mean gradients of the last ten data epochs, the optimiser buffers and the validation history by running Python code in a restricted sandbox, and it may score a few candidate models on the validation set. It then submits a short program that computes new weights from this history, together with a plan of ten letters that marks each of the next ten epochs as a data epoch (D) or a virtual epoch (V). It may also set the number of epochs after which training should end. At a V slot the training loop calls the program instead of passing over the training data.

Every accepted virtual epoch takes the place of a data epoch, at the cost of a few forward passes over 5{,}000 validation images instead of a forward and backward pass over 45{,}000 training images. An acceptance rule protects accuracy at each step. A proposed state is kept only if its validation accuracy and loss stay within a small tolerance of the last accepted state. Otherwise the model returns to that state, and the slot and the rest of the window run data epochs. A program that produces bad states therefore loses at most the validation passes of one slot per window, although a sequence of accepted states can still leave a run behind the recipe, as Section~\ref{sec:results} shows. Because the program sees the whole recent history, it can replay gradients, extrapolate the trajectory, average snapshots or treat each layer differently, and the LLM chooses among such updates from the evidence of the current run rather than from a rule fixed before training. The LLM is called at most seven times in an 80-epoch run, so the API cost is bounded and does not grow with the size of the model. Every decision is stored as code, together with the rationale the LLM gave for it.

Earlier work skips training by predicting future weights with a predictor fixed before the run~\cite{sinha2017introspection,jang2023wnn,knyazev2025nino,shou2026farcasting,mcentire2026leap}, or meta-trains optimisers across many runs and then freezes them~\cite{andrychowicz2016learning,metz2022velo,chen2023lion}. The closest system, OPAL~\cite{bai2026reactive}, lets a Claude Opus~5.5 agent rewrite the update rule during a run, but every one of its steps still uses data. \method{} asks instead whether LLM-written programs can stand in for data epochs of a recipe that already works. Claims of this kind are easy to overstate, because the same recipe run for fewer epochs, with its schedule decayed to zero, is a strong competitor~\cite{li2020budgeted,kaddour2023notrain}. We therefore compare every run with the standard recipe trained for the same number of data epochs, as well as with the full-length recipe. To separate what the LLM contributes from what its acceptance rule and plans contribute, we also run two fixed controls with the same machinery: one whose update candidates come from prior work, and one that encodes the update the agents found.

We make four contributions.
\begin{itemize}
  \item We formulate the replacement of data epochs by training-data-free virtual epochs on a fixed schedule clock, with a validation acceptance rule, a window plan and an optional earlier end. We evaluate it against the accuracy-versus-data frontier of the standard recipe (Section~\ref{sec:problem}).
  \item We present \method{}, in which a tool-using LLM analyses the weight and gradient history in a sandbox and writes the virtual-update program, the plan and the schedule horizon. Programs never compute gradients on data, every virtual epoch is checked on the validation set, and every API call is logged (Section~\ref{sec:method}).
  \item We evaluate \method{} with Claude Opus~5.5~\cite{anthropic2026opus55} and Claude Sonnet~5~\cite{anthropic2026sonnet5} in 120 runs over three architectures and two datasets. Opus used 32.8\% fewer data epochs than the 80-epoch recipe at 0.31 percentage points lower accuracy (95\% CI $-0.72$ to $+0.09$; up to 0.94 points on ResNet-18), which places it on or slightly above the recipe's frontier (+0.18 points, not distinguishable from zero). Two fixed controls with the same machinery show that the savings come from the update the LLM chose: with candidates from prior work a rule saved 5.4\% of the data epochs, and with the agents' averaging added, 25.0\% (Sections~\ref{sec:setup} and~\ref{sec:results}).
  \item We analyse all 167 programs and every accepted virtual epoch. Without being prompted, Opus averaged recent weight snapshots in every run and interleaved the averages with data epochs, whereas most of Sonnet's virtual epochs barely moved the weights. We also report GPU time, API cost and latency (Sections~\ref{sec:results} and~\ref{sec:discussion}).
\end{itemize}
Section~\ref{sec:related} reviews related work, Section~\ref{sec:problem} states the problem, Section~\ref{sec:method} describes \method{}, Sections~\ref{sec:setup} and~\ref{sec:results} report the experiments, Section~\ref{sec:discussion} discusses limitations and Section~\ref{sec:conclusion} concludes.

\section{Related work}
\label{sec:related}

Several methods accelerate training by predicting weights instead of computing more gradient steps. Introspection trains a network on the weight histories of one task to predict the value a weight will reach many steps later, and uses it to jump ahead periodically when other networks are trained~\cite{sinha2017introspection}. WNN periodically nowcasts near-future weights with a learned predictor in order to skip epochs~\cite{jang2023wnn}. NiNo builds the nowcaster as a graph network over the neuron connectivity of the model and reports that it accelerates Adam by up to 50\%~\cite{knyazev2025nino}, and farcasting predicts weights far ahead with a one-layer network~\cite{shou2026farcasting}. Leap+Verify extrapolates the trajectory analytically, with linear, quadratic or momentum-based formulas, and accepts a prediction only if it passes a check on held-out loss~\cite{mcentire2026leap}. That check is the closest prior counterpart of our acceptance rule. In all of these methods the predictor, or the family of predictors and the rule that gates them, is fixed before the run starts. In \method{} the predictor is a program that an LLM writes during the run and may replace every ten epochs, and it may compute anything from the recorded history as long as it touches no training data. Data pruning saves work in another way, by reducing the number of training examples~\cite{sorscher2022beyond}; \method{} keeps every example and reduces the number of passes over them.

Weight averaging along the trajectory is relevant here because Opus, and Sonnet in one run, chose it in our experiments. Stochastic weight averaging (SWA) averages the points that SGD visits with a cyclical or constant learning rate, recomputes the batch-normalisation statistics~\cite{ioffe2015batch} of the average, and finds wider optima that generalise better~\cite{izmailov2018swa}. LAWA averages the latest $k$ end-of-epoch checkpoints during training and reaches intermediate accuracies dozens of epochs earlier on ImageNet and in language-model pre-training, although its head start shrinks towards the end of training~\cite{kaddour2022lawa}. In both methods the average is kept apart from the SGD trajectory, which continues from its own weights. Lookahead runs $k$ steps of an inner optimiser, moves a set of slow weights part of the way towards the result and restarts the inner optimiser from them~\cite{zhang2019lookahead}. The programs Opus wrote compute LAWA-like means, mostly of the last three to five snapshots, but an accepted mean replaces the weights and training continues from it. Together with the alternating plans of Opus, which interleave such means with data epochs, this resembles Lookahead more than SWA. Our hand-designed control is a LAWA-like rule with fixed timing whose averages also replace the weights. These methods were published well before the training cut-offs of both LLMs~\cite{anthropic2026opus55docs,anthropic2026sonnet5docs}, so we describe what Opus did as an unprompted choice of a known technique, not as an invention.

Learned optimisers replace the update formula by a network meta-trained across tasks, from LSTM optimisers trained by gradient descent~\cite{andrychowicz2016learning} to VeLO, meta-trained with about 4{,}000 TPU-months of compute~\cite{metz2022velo}. Lion was found by a symbolic program search over many training runs~\cite{chen2023lion}. All three are frozen once the target model starts training. LLMs have also served as optimisers that propose new candidates from a history of scored ones~\cite{yang2024opro}. FunSearch pairs LLM-written programs with an automatic evaluator~\cite{romeraparedes2024funsearch}, and \method{} follows the same pattern, with the validation set as evaluator and programs that act on weights. OPAL is the closest system~\cite{bai2026reactive}. Its agent, Claude Opus~5.5, watches a single run and keeps rewriting two programs while training continues: one that probes the training dynamics and one that defines the update rule of each parameter group. The agent sees the statistics its probes return rather than raw weights or gradients, starts from scratch in every run, and targets a better final model in language modelling and image generation, where no tuned recipe exists. In a ResNet-18 experiment on CIFAR-10 with noisy labels, OPAL's agent also chose weight averaging selected on validation data, so an LLM agent choosing averaging has been reported before. \method{} differs in what the LLM sees and in what its programs are for: the LLM reads weight snapshots and epoch gradients directly in a sandbox, and its programs replace data epochs of an established recipe. In earlier work we applied the same pattern to federated learning: in FedReason an LLM inspects client models every ten rounds in a server-side sandbox and writes the aggregation program~\cite{ruzbahani2026fedreasonlargelanguagemodels}. \method{} moves this loop from aggregation across clients to the training trajectory of a single model. Takehi et al.\ reported, in a blog post that has not been peer reviewed, that a frontier LLM can tell properties of small networks from their weights and edit them without gradient training~\cite{takehi2026sparks}. The post anticipates that LLMs will optimise weights by reasoning about them, and our experiments test a narrow version of that expectation inside a training run.

Related research has examined complementary challenges in
the adaptation and maintenance of learned models.
In ~\cite{ruzbahani2026hqrl}, we can find the
investigated hierarchical quantum reinforcement learning
for defending federated learning against multi-vector
poisoning attacks. In another direction, authors in \cite{a2} examined federated
unlearning in agentic energy systems, emphasizing
security and governance implications. While these
studies focus on adversarial robustness and model
unlearning rather than training acceleration, they
illustrate the broader importance of controlling
model updates in distributed and adaptive learning
systems. GradReason addresses a different question:
whether an LLM can construct weight-space updates
during an individual training run that replace
data-intensive optimization steps while preserving
predictive performance.

Claims of faster training depend on the reference. Budgeted training showed that, for a fixed number of iterations, a learning-rate schedule that decays to zero at the end of the budget is preferable to a longer schedule cut short~\cite{li2020budgeted}. Kaddour et al.\ found that several efficient-training methods for transformer language models (BERT and T5 pre-training) lose their advantage against such budget-matched baselines~\cite{kaddour2023notrain}. We follow this advice: every budget of our baseline has its own cosine schedule that ends at zero, and the main comparison is with the baseline at the same number of data epochs.

\section{Problem formulation}
\label{sec:problem}

We first fix the recipe and its notation, then define the schedule clock, virtual epochs and the decisions of the meta-learner, and finally the measures by which we judge it. Let $\mathcal{D}$ be a training set of $N$ labelled images, $\mathcal{V}$ a disjoint validation set and $\mathcal{T}$ a test set. A model $f(\cdot\,;w)$ with parameters $w$ is trained by a standard recipe, which consists of an optimiser $\mathcal{O}$ with state $m$ (momentum or Adam moments) and a learning-rate schedule $\eta_B(t)$. The schedule warms up linearly and then decays along a cosine to zero at $t=B$, where $B$ is the epoch budget. One data epoch applies $S$ optimiser steps over a shuffled pass of $\mathcal{D}$,
\begin{equation}
  \begin{split}
  (w_e,m_e)&=\mathcal{O}\big(w_{e-1},m_{e-1};\mathcal{D},\eta\big),\\
  G_e&=\frac{1}{S}\sum_{k=1}^{S}\nabla_w\ell_k\big(w_{e,k}\big),
  \end{split}
  \label{eq:epoch}
\end{equation}
where $\ell_k$ is the mini-batch loss of step $k$, $w_{e,k}$ the weights before that step and $G_e$ the epoch-mean gradient, which we record. We write $a_s(B)$ for the test accuracy of the recipe with budget $B$ and seed $s$. A seed fixes the split of the data, the initial weights and the order of the mini-batches in every epoch.

The methods we study run on a schedule clock of $E=80$ epochs. Until the horizon is changed, clock epoch $e$ has the learning rate of the recipe with $B=E$, whatever happens in that epoch, and it is either a data epoch or a virtual epoch. A virtual epoch applies a program $\pi$ to the history
\begin{equation}
  \mathcal{H}_{e}=\big(\{W_{e'}\}_{e-10<e'\le e},\ \{G_{e'}\}_{e'\in\mathcal{E}_e},\ m_e,\ \mathcal{R}_e\big),
  \label{eq:history}
\end{equation}
which holds the last ten weight snapshots (from data epochs and accepted virtual epochs) and the epoch-mean gradients of Eq.~\eqref{eq:epoch} for the last ten data epochs $\mathcal{E}_e$. It also holds the current optimiser state and the records $\mathcal{R}_e$ of validation metrics, learning rates and gradient statistics. The program returns a candidate $\tilde w=\pi(\mathcal{H}_{e-1})$ without touching $\mathcal{D}$. It may call a forward evaluation $\mathrm{ev}_{\mathcal{V}}(w)$, which returns the validation accuracy $A_{\mathcal{V}}(w)$, the loss $L_{\mathcal{V}}(w)$ and the per-class accuracy, at most $\bprog$ times. For a model with batch normalisation, every evaluation first recomputes the normalisation statistics on validation images. The candidate is accepted if
\begin{equation}
  \begin{split}
  &A_{\mathcal{V}}(\tilde w)\ \ge\ A_{\mathcal{V}}(w_{\mathrm{ref}})-\delta_A
  \quad\text{and}\\
  &L_{\mathcal{V}}(\tilde w)\ \le\ (1+\delta_L)\,L_{\mathcal{V}}(w_{\mathrm{ref}}),
  \end{split}
  \label{eq:accept}
\end{equation}
where $w_{\mathrm{ref}}$ is the most recent state that a data epoch produced or a virtual epoch had accepted, $\delta_A=0.3$ percentage points (pp) and $\delta_L=0.01$. An accepted candidate becomes $w_e$, and the optimiser state stays as it was unless the program also replaced it. A rejected candidate is discarded, and the slot runs a data epoch from $w_{\mathrm{ref}}$ instead.

A meta-learner $\mathcal{M}$ decides how the clock is used. At the analysis points $c_j=10j$, $j=1,\dots,7$, that lie before the end of the run, it observes
\begin{equation}
  \mathcal{I}_j=\big(\mathcal{H}_{c_j},\ \nu_{j-1},\ \pi_{j-1},\ \omega_{j-1}\big),
\end{equation}
that is, the history, its own notes $\nu_{j-1}$, the previous program and the outcome $\omega_{j-1}$ of the previous window (which slots were accepted or failed, and why). It outputs a decision $(\pi_j,p_j,f_j)$: a program, a window plan $p_j\in\{\mathrm{D},\mathrm{V}\}^{n_j}$ with one letter for each of the $n_j=\min(10,E'-c_j)$ clock epochs of the coming window, and an optional horizon $f_j$. The plan is executed in order, and after the first V slot whose candidate fails, the remaining V slots of the window become D. A horizon $f_j$ moves the final epoch to $E'=c_j+f_j$ and re-shapes the learning rate of the remaining epochs into a half-cosine from its current value to zero at $E'$. It can only shorten the run, and the run cannot end before epoch 20. Epochs 1 to 10 are always data epochs.

With $D$ data epochs, $V$ accepted virtual epochs and the final epoch $E'\le E$, we have $D+V=E'$, and the share of data epochs saved is $1-D/E$. The goal is the smallest $D$ that keeps the accuracy of the full recipe,
\begin{equation}
  \begin{split}
  &\min_{\mathcal{M}}\ \mathbb{E}[D]
  \quad\text{s.t.}\quad
  \mathbb{E}\big[a(w_{E'})\big]\ \ge\ a_s(E)-\epsilon,\\
  &e^{\mathrm{inv}}_j\le\binv\ \ \forall j,\qquad
  e^{\mathrm{prog}}_e\le\bprog\ \ \forall e,
  \end{split}
  \label{eq:meta}
\end{equation}
where $\epsilon$ is of the order of the variation between seeds. The budgets bound the validation evaluations made during invocation $j$ ($e^{\mathrm{inv}}_j$) and by the program at clock epoch $e$ ($e^{\mathrm{prog}}_e$). The expectation is over seeds and over the randomness of $\mathcal{M}$. Two further constraints are structural: neither a program nor the analysis may compute a gradient on any data, and the test set is never observed. A meta-learner that always plans D and never changes the horizon reproduces the 80-epoch recipe.

Shortening the run always saves data epochs, but it only moves a meta-learner along the recipe's own trade-off between data and accuracy. We therefore measure what the meta-learner adds relative to that trade-off. For seed $s$, the baseline frontier $F_s(D)$ interpolates $a_s(B)$ at $B\in\{10,20,30,40,60,80\}$ linearly in $\log B$, and we report the gaps
\begin{equation}
  \Delta_{80}=a(w_{E'})-a_s(80),\qquad \Delta_F=a(w_{E'})-F_s(D).
  \label{eq:gaps}
\end{equation}
A method with $\Delta_F\le0$ saved nothing that the recipe with a shorter budget would not also have saved.

\section{The \method{} framework}
\label{sec:method}

\method{} is a meta-learner for problem~\eqref{eq:meta} in which an LLM makes every decision (Fig.~\ref{fig:overview} and Algorithm~\ref{alg:method}). The run starts with ten data epochs that fill the history. At each analysis point the training loop opens a new tool-use conversation with the LLM, loads the history into an analysis sandbox and waits until the LLM submits a decision that passes a dry run. The plan of that decision is executed over the next ten epochs, the program is called at the V slots, and no API call is made until the next analysis point. A window without an accepted submission runs ten data epochs; in our runs this never happened.

\begin{algorithm}[!t]
\caption{\method{}: one training run.}
\label{alg:method}
\begin{algorithmic}
\Require recipe $\mathcal{O}$, clock $E$, period $T$, budgets $\binv$ and $\bprog$, tolerances $\delta_A$ and $\delta_L$
\State run data epochs $1,\dots,T$ and record $W_e$, $G_e$ and the validation metrics;\ \ $E'\gets E$;\ \ $\nu\gets\varnothing$;\ \ $\omega\gets\varnothing$
\For{$c=T,2T,\dots$ while $c<E'$}
  \State $(\pi,p,f,\nu)\gets\textsc{Invoke}(\mathcal{H}_c,\nu,\omega)$
  \If{$f\neq\varnothing$}
    \State $E'\gets c+f$; re-shape the learning rate of epochs $c+1,\dots,E'$
  \EndIf
  \State $w_{\mathrm{ref}}\gets w_c$;\ \ $\mathit{failed}\gets$ false
  \For{$k=1,\dots,|p|$, with $e=c+k$}
    \State $\mathit{accepted}\gets$ false
    \If{$p_k=\mathrm{V}$ and not $\mathit{failed}$}
      \State $\tilde w\gets\pi(\mathcal{H}_{e-1})$ \Comment{no training data; at most $\bprog$ evaluations}
      \State $\mathit{accepted}\gets\tilde w$ is valid and satisfies Eq.~\eqref{eq:accept};\ \ $\mathit{failed}\gets$ not $\mathit{accepted}$
    \EndIf
    \If{$\mathit{accepted}$}
      \State $w_e\gets\tilde w$; add the snapshot $W_e=w_e$ to the history
    \Else
      \State run data epoch $e$ from $w_{\mathrm{ref}}$; add $W_e$ and $G_e$ to the history
    \EndIf
    \State $w_{\mathrm{ref}}\gets w_e$
  \EndFor
  \State $\omega\gets$ outcome of every slot of the window
\EndFor
\State \Return $w_{E'}$
\Statex
\Function{Invoke}{$\mathcal{H}_c,\nu,\omega$}
  \State load $\mathcal{H}_c$ into the sandbox; send the brief built from $\mathcal{H}_c$, $\nu$ and $\omega$
  \For{at most ten model requests}
    \If{the LLM runs analysis code in the sandbox}
      \State run the code and return its printed output \Comment{at most $\binv$ evaluations in total}
    \ElsIf{the LLM submits $(\pi',p',f',\nu')$}
      \If{$p'$ and $f'$ are admissible and $\pi'$ passes a dry run}
        \State \Return $(\pi',p',f',\nu')$
      \EndIf
      \State return the error to the LLM \Comment{at most three submissions}
    \EndIf
  \EndFor
  \State \Return (no program, $\mathrm{D}\cdots\mathrm{D}$, $\varnothing$, $\nu$)
\EndFunction
\end{algorithmic}
\end{algorithm}

\subsection{History and sandbox}
\label{sec:sandbox}

During an invocation the LLM has two tools: one runs analysis code in a persistent sandbox and returns what it prints, and the other submits a decision. The sandbox holds copies of the history of Eq.~\eqref{eq:history}: the current weights, the weight snapshots of the last ten epochs, the epoch-mean gradients of the last ten data epochs, the optimiser buffers and the per-epoch records. It also holds a description of the setting, with the architecture, its layer groups, the recipe, the learning rate of every remaining epoch, the current horizon and the budgets. All tensors are float32 copies on the CPU. Helpers compute cosines between states, a principal component analysis of the snapshot trajectory and the learning rate at any point of the clock, and let a draft program be tried before it is submitted. A budgeted evaluator scores any state on the validation set, at most $\binv=8$ times per invocation.

A summary table gives the LLM a starting point. It is organised by layer group, for example the stem, the four stages and the head of ResNet-18, or the embedding, three groups of four transformer blocks and the head of ViT-Tiny. For each group and recent epoch it lists the norm of $G_e$, the cosine between consecutive epoch gradients, the relative weight change $\lVert W_e-W_{e-1}\rVert/\lVert W_{e-1}\rVert$ and the alignment of that change with $-G_e$. Per group it adds the trend of the gradient norm, the coherence of the epoch gradients over ten epochs, how far the optimiser moved per plain gradient step and a secant estimate of curvature times step size. Further columns give the per-step signal-to-noise ratio of the gradients, the cosine between the momentum buffer and the latest gradient, and the correlation between update size and the decrease of the validation loss. The table ends with the validation history and a short reading guide.

The training images, the test set and the model object are never placed in the sandbox, so no code can compute a gradient on data. Code runs with automatic differentiation disabled, a restricted set of built-in functions and a whitelist of numerical libraries. Before it runs, a static check rejects access to files, processes, private attributes and optimisers, and while it runs, file writes, network access and subprocesses are blocked. Printed output is truncated to 10{,}000 characters.

\subsection{Programs, plans and the horizon}
\label{sec:programs}

A program is a Python function that the training loop calls at every V slot. It receives the history as it is at the slot, together with the clock epoch, the position of the slot in the plan, the learning rate of the epoch and a validation evaluator limited to $\bprog=3$ calls per slot. A program returns a complete state with the keys and shapes of the current one, or it declines the slot with a reason, which fails the slot and is reported at the next analysis point. It may also replace the optimiser buffers, and the replacement takes effect only if its state is accepted. A call must finish within 3~s, not counting validation. Snapshots of accepted virtual epochs enter the history, but its gradients come from data epochs only, so after several virtual epochs the newest gradient is older than the weights.

A submission consists of the program, a rationale of two to six sentences, notes of at most 4{,}000 characters for the next invocation, the plan and the optional horizon. The training loop checks the length and letters of the plan and the admissible range of the horizon, loads the program and, if the plan contains a V, dry-runs the program once for the first V slot. Errors go back to the LLM, which may resubmit, up to three submissions per invocation, and an accepted submission ends the invocation. Each invocation is a new conversation of at most ten model requests. The LLM keeps no memory between invocations; what it knows about earlier windows comes from the first message of each conversation, which we call the brief. The brief gives the setting, the data accounting and the validation history. It reports the outcome of each slot of the last window with rejection margins and the program's own diagnostics, and it repeats the previous program with its rationale, the LLM's own notes and the budgets. Notes never pass from one run to another.

The system prompt is static. It describes the clock, the two levers and their accounting, and the interfaces. Its six rules forbid computing gradients on data, reading test data and accessing files, the network or other processes. They also ask for cheap and numerically safe programs, restrict validation to choosing among a few structurally different candidates, and require an accepted submission at the end of every invocation. The prompt states the goal as matching the full recipe within validation noise with as few data epochs as possible, and it says that trading accuracy for saved epochs counts as a failure. It announces that the run will be judged against the recipe trained with the same number of data epochs, and it warns that a candidate which barely moves the weights passes the acceptance rule without making the progress of a data epoch. A section on the evidence explains what a data epoch does under SGD and AdamW, what consistent and noise-dominated epoch gradients look like, how curvature and the age of the evidence can be read, and how batch normalisation and weight decay interact. The prompt names no update technique and recommends none, and it never mentions averaging of weights. It does say that an epoch-mean gradient is dominated by mini-batch noise when consecutive ones are uncorrelated, and the principal component helper returns, among other things, the mean of the snapshots it analyses, so everything needed to compute an average was at hand. Its only code is a skeleton of the program contract that copies the current weights.

\subsection{Safeguards and cost}
\label{sec:safeguards}

The acceptance rule~\eqref{eq:accept} is the main safeguard. It compares every candidate with the state the candidate would replace, so a program cannot lose much accuracy within a single slot, and a failure costs one slot, after which the window runs data epochs. The rule cannot see what the replaced data epoch would have achieved. A candidate that leaves the weights almost unchanged passes it, consumes a clock epoch of the decaying schedule and counts as a saved data epoch. The prompt says so, and the frontier gap of Eq.~\eqref{eq:gaps} charges such epochs at their true value. Code written by an LLM can be faulty, so the training loop treats it as untrusted. It checks that every returned state has the right keys, shapes and finite values, enforces the time limit and the evaluation budgets, and keeps test metrics out of every record that a program or the LLM can read. Section~\ref{sec:discussion} discusses the limits of this sandbox.

Every API call is logged with its token counts, the model that served it and its latency. Its cost is
\begin{equation}
  c=10^{-6}\big(n_{\mathrm{in}}\,p_{\mathrm{in}}+n_{\mathrm{cw}}\,p_{\mathrm{cw}}+n_{\mathrm{cr}}\,p_{\mathrm{cr}}+n_{\mathrm{out}}\,p_{\mathrm{out}}\big)\ \text{USD},
  \label{eq:cost}
\end{equation}
where $n_{\mathrm{in}}$, $n_{\mathrm{cw}}$, $n_{\mathrm{cr}}$ and $n_{\mathrm{out}}$ count the uncached input, cache-write, cache-read and output tokens, thinking included. The list prices per million tokens $p_{\mathrm{in}}$, $p_{\mathrm{cw}}$, $p_{\mathrm{cr}}$ and $p_{\mathrm{out}}$ are 4, 5, 0.20 and 20~USD for Claude Opus~5.5 and 2, 2.50, 0.20 and 10~USD for Claude Sonnet~5~\cite{anthropic2026opus55docs,anthropic2026sonnet5docs}. The system prompt and the tool definitions are identical across calls and are served from the prompt cache. A run has at most seven invocations of at most ten requests each, so by Eq.~\eqref{eq:cost} its API cost depends neither on the size of the model nor on the size of the dataset. A global budget stops all API calls once the total spend of the study reaches it.

The method also costs GPU time. Data epochs of \method{} accumulate $G_e$, which made them 4\% to 10\% slower than the recipe's in five of our six settings. Every slot, accepted or not, also needs validation passes and, for models with batch normalisation, a recalibration of the statistics on 5{,}000 images. In our implementation training waits while the LLM works, so its latency adds to the wall-clock time. Section~\ref{sec:res-cost} measures both.

\section{Experimental setup}
\label{sec:setup}

We use CIFAR-10 and CIFAR-100~\cite{krizhevsky2009learning}. For each seed, a class-balanced validation set of 5{,}000 images is drawn from the 50{,}000 training images, which leaves 45{,}000 for training, and the 10{,}000 test images are used for reporting only. Training images are augmented on the GPU with random crops after 4-pixel zero padding and with horizontal flips. Batch-normalisation statistics are recalibrated on the validation images under one fixed draw of this augmentation, because clean images shift the statistics. Table~\ref{tab:setup} lists the three architectures. ResNet-18~\cite{he2016deep} and MobileNetV2~\cite{sandler2018mobilenetv2} are adapted to $32\times32$ inputs with a stride-1 $3\times3$ stem and no max-pooling (MobileNetV2 also uses stride 1 in its second stage), and both use batch normalisation~\cite{ioffe2015batch}. ViT-Tiny~\cite{dosovitskiy2021vit} has the width of DeiT-Tiny~\cite{touvron2021deit}: $4\times4$ patches, width 192 and 12 pre-norm blocks with three heads, a class token and learned position embeddings.

The recipes follow common practice. Both convolutional networks use SGD with Nesterov momentum 0.9, learning rate 0.1 at batch size 128, weight decay $5\times10^{-4}$ and a one-epoch linear warm-up followed by cosine decay to zero~\cite{loshchilov2017sgdr}. ViT-Tiny uses AdamW~\cite{loshchilov2019adamw} with learning rate $10^{-3}$ at batch size 128 and weight decay 0.05, which is not applied to normalisation parameters, biases, the class token and the position embeddings. Its five-epoch warm-up, cosine decay, weight decay and label smoothing of 0.1~\cite{szegedy2016rethinking} follow the DeiT recipe~\cite{touvron2021deit}. Everything is implemented in PyTorch~\cite{paszke2019pytorch} with bfloat16 autocasting. Each run holds its images in GPU memory on a single NVIDIA RTX~5090, which several runs shared at a time.

\begin{table*}[t]
\centering
\caption{Architectures, recipes and test accuracy of the 80-epoch recipe (mean\,$\pm$\,s.d., two seeds).}
\label{tab:setup}
\small
\setlength{\tabcolsep}{4pt}
\begin{tabular}{@{}llllcc@{}}
\toprule
 & Parameters (M) & & Data epoch (s) & \multicolumn{2}{c}{Accuracy, $B=80$ (\%)} \\
\cmidrule(lr){5-6}
Architecture & CIFAR-10 / 100 & Recipe & CIFAR-10 / 100 & CIFAR-10 & CIFAR-100 \\
\midrule
ResNet-18 & 11.17 / 11.22 & SGD, lr 0.1, wd $5\times10^{-4}$, warm-up 1 & 2.4 / 2.0 & 94.47\pmstd{0.05} & 76.25\pmstd{0.23} \\
MobileNetV2 & 2.24 / 2.35 & SGD, lr 0.1, wd $5\times10^{-4}$, warm-up 1 & 5.6 / 5.9 & 92.73\pmstd{0.25} & 72.20\pmstd{0.42} \\
ViT-Tiny & 5.36 / 5.38 & AdamW, lr $10^{-3}$, wd 0.05, warm-up 5, LS 0.1 & 5.8 / 5.8 & 83.15\pmstd{0.15} & 55.17\pmstd{0.81} \\
\bottomrule
\end{tabular}
\end{table*}

We compare five methods in all six settings, each with seeds 0 and 1. The standard recipe runs at budgets $B\in\{10,20,30,40,60,80\}$, each with its cosine fitted to $B$. Its 72 runs trace the frontier $F_s$ and give the 80-epoch reference. \method{} runs on the clock of $E=80$ epochs with Claude Opus~5.5~\cite{anthropic2026opus55} or Claude Sonnet~5~\cite{anthropic2026sonnet5} as the LLM, both with adaptive thinking and medium reasoning effort. Medium is the default effort of Opus~5.5, whereas Sonnet~5 defaults to high effort, so we set it explicitly; neither model accepts a temperature or other sampling parameter~\cite{anthropic2026opus55docs,anthropic2026sonnet5docs}. We call the two configurations Opus and Sonnet; the figure legends label them LLM agent (Opus~5.5) and LLM agent (Sonnet~5). The control, which we call the rule, uses the same clock, levers, acceptance rule and validation budget, but its decisions are fixed. Its plan is DDDDDDDDVV in every window. Its program has four candidates, the means of the last three and of the last five data-epoch snapshots and the extrapolations $W_e+\alpha(W_e-W_{e-1})$ with $\alpha\in\{0.5,1\}$. Within its three evaluations it scores the shorter mean and the smaller step first, tries the longer mean or the larger step only if the shorter one lowered the validation loss, spends any evaluation left on candidates not yet tried, and returns the best. At analysis points from epoch 40 on, the rule ends the run ten epochs later if the best validation accuracy rose by less than 0.2~pp over the last window. We designed the rule after a pilot in which both LLMs had averaged snapshots (Section~\ref{sec:res-behaviour}), so it encodes their behaviour rather than an independent idea of ours. A second control, the \emph{a-priori rule}, is identical to the rule, including its plan, early finish and budgets, except that its candidates are only the two extrapolations. It is the rule a designer would write from the closest prior method, Leap+Verify~\cite{mcentire2026leap}, without knowing what the agents would find, and the pair of rules isolates the contribution of the update from that of the clock, the plans and the acceptance rule. The study has 120 runs: 72 of the recipe and 12 each of the two rules, Opus and Sonnet.

We report the test accuracy of the model after the last clock epoch, the data epochs $D$, the share saved $1-D/80$, the gaps $\Delta_{80}$ and $\Delta_F$ of Eq.~\eqref{eq:gaps}, GPU time and API cost. For a model with batch normalisation whose last epoch was virtual, the final model carries the statistics recomputed on validation images, and we repeated the comparisons with BN-recalibrated test accuracy on both sides (Section~\ref{sec:res-frontier}). Every gap is paired with the baseline runs of the same setting and seed, which share the split, the initial weights and the mini-batches of every data epoch. The frontier of a run interpolates the six baseline budgets of its own seed. Pooled statistics use the 12 paired differences per method: the mean, the sample standard deviation, a 95\% $t$ interval, wins and losses, the exact two-sided Wilcoxon signed-rank test and the sign test. With 12 pairs these tests detect only large effects, and we treat them as indicative. GPU time is not the measured wall time, because several runs shared the GPU. We rebuilt it from the data epochs, validation passes, recalibrations and program calls that each run logged, with per-setting times measured on the otherwise idle GPU (five repetitions each). The LLM's latency is not GPU time and is reported separately.

\section{Results}
\label{sec:results}

All 120 runs finished. Each of the 167 invocations ended with a submission that passed its dry run at the first attempt. The requested model served every API call, and no request was refused. Table~\ref{tab:pooled} summarises the 12 runs of each method and Table~\ref{tab:settings} breaks them down by setting. Fig.~\ref{fig:frontier} places the runs on the frontier, Figs.~\ref{fig:timeline} and~\ref{fig:curves} show what the methods did during training, and Fig.~\ref{fig:cost} shows the cost.

\subsection{Data epochs saved and the baseline frontier}
\label{sec:res-frontier}

\begin{table*}[t]
\centering
\caption{Pooled results, 12 runs per method, paired by setting and seed ($\pm$: s.d.; $p$: Wilcoxon).}
\label{tab:pooled}
\small
\setlength{\tabcolsep}{3.5pt}
\begin{tabular}{@{}lccccccccc@{}}
\toprule
Method & $D$ & Saved (\%) & GPU saved (\%) & $\Delta_{80}$ (pp) [95\% CI] & W/L & $p$ & $\Delta_F$ (pp) [95\% CI] & W/L & $p$ \\
\midrule
A-priori rule & 75.7 & 5.4\pmstd{10.1} & 0.9\pmstd{11.4} & $+0.02$\ci{-0.17}{+0.22} & 3/7 & 0.34 & $+0.13$\ci{-0.22}{+0.47} & 5/7 & 0.97 \\
Rule & 60.0 & 25.0\pmstd{12.2} & 19.6\pmstd{14.2} & $-0.19$\ci{-0.43}{+0.04} & 3/9 & 0.12 & $+0.29$\ci{-0.01}{+0.60} & 8/4 & 0.09 \\
Sonnet & 63.8 & 20.3\pmstd{10.2} & 12.1\pmstd{11.2} & $-0.17$\ci{-0.33}{-0.01} & 4/8 & 0.08 & $+0.15$\ci{-0.21}{+0.51} & 7/5 & 0.68 \\
Opus & 53.8 & 32.8\pmstd{7.3} & 22.5\pmstd{8.8} & $-0.31$\ci{-0.72}{+0.09} & 3/9 & 0.11 & $+0.18$\ci{-0.18}{+0.54} & 7/5 & 0.52 \\
\bottomrule
\end{tabular}
\end{table*}

\begin{figure*}[t]
  \centering
  \includegraphics[width=\textwidth]{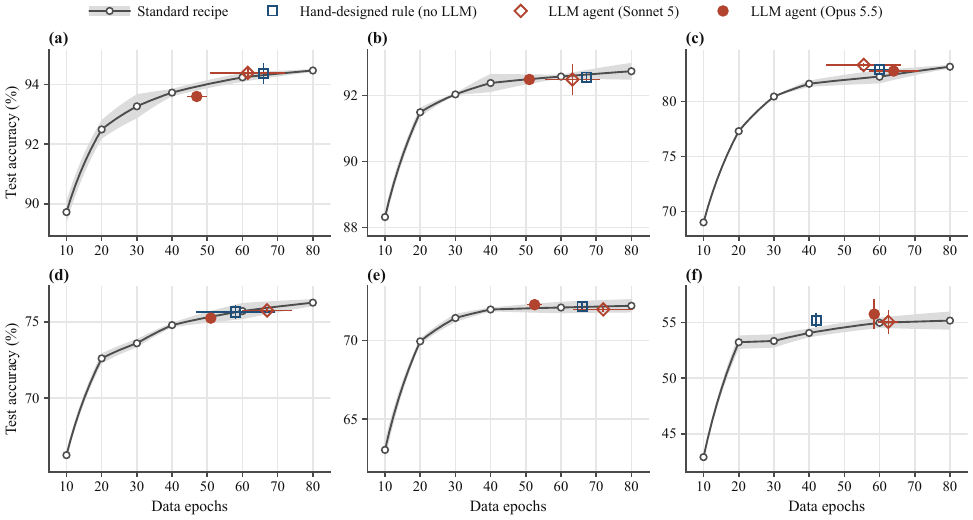}
  \caption{Test accuracy against data epochs. Rows: CIFAR-10, CIFAR-100; columns: ResNet-18, MobileNetV2, ViT-Tiny.}
  \label{fig:frontier}
\end{figure*}

With Opus, \method{} used 53.8 data epochs on average instead of 80, a saving of 32.8\% (standard deviation 7.3), and its final test accuracy was 0.31~pp below that of the 80-epoch recipe (95\% CI $-0.72$ to $+0.09$; 3 wins, 9 losses; Table~\ref{tab:pooled}). Sonnet saved 20.3\% of the data epochs at a loss of 0.17~pp and the rule 25.0\% at a loss of 0.19~pp. Only Sonnet's interval excludes zero; its Wilcoxon test does not ($p=0.077$), although it does with BN-recalibrated accuracy on both sides ($p=0.028$). Part of these savings came from moving along the recipe's own trade-off between data and accuracy. Against the frontier at the same number of data epochs, Opus was 0.18~pp more accurate (95\% CI $-0.18$ to $+0.54$; 7 wins, 5 losses; $p=0.52$), Sonnet 0.15~pp ($p=0.68$) and the rule 0.29~pp ($-0.01$ to $+0.60$; 8 wins, 4 losses; $p=0.09$). None of the three is distinguishable from the frontier in 12 runs. This does not change against the seed-mean frontier that Fig.~\ref{fig:frontier} draws (+0.16, +0.15 and +0.27~pp) or with BN-recalibrated test accuracy on both sides (+0.17, +0.15 and +0.28~pp).

Counting data epochs to a target accuracy gives the same result. We counted the data epochs a run needed to come within 0.5~pp of the final accuracy of the 80-epoch recipe of its seed, with BN-recalibrated test accuracy for the target and for every epoch. The 80-epoch recipe got there after 66.4 data epochs on average. Opus got there in 8 of 12 runs, after 39.1 data epochs on average, and stayed there until the end in 6; Sonnet and the rule got there in 12 and 11 runs, after 56.7 and 44.9 data epochs. We also chose, in hindsight and per seed, the shortest baseline budget that reached the target. Relative to that budget no method was faster by a margin that 12 runs can resolve (Opus minus the best budget: $-9.1\pm18.4$ data epochs at the first crossing, $+4.2\pm12.7$ when the target had to hold to the end).

The architectures behaved differently (Table~\ref{tab:settings} and Fig.~\ref{fig:frontier}, whose bars and band show one standard deviation over the two seeds; points that would hide each other are shifted sideways by at most 2 data epochs). On ResNet-18 and MobileNetV2 the three methods lie on the frontier: they reach fewer data epochs by moving along it, not above it. Opus went furthest on ResNet-18, where it saved 38.8\% of the data epochs and ended 0.94~pp below the 80-epoch recipe, with a frontier gap of $-0.20$~pp (1 win in 4). That loss is three times the acceptance tolerance and falls short of the goal stated in the prompt, although each of its virtual epochs had passed the step-wise acceptance check. Most of it is the price of the shorter budget rather than of the virtual epochs: at the 49 data epochs Opus used, the recipe's own frontier lies 0.74~pp below its 80-epoch accuracy, and Opus ended 0.20~pp below that frontier. Opus was the most aggressive on ResNet-18 (4.4 virtual epochs per window, against 4.0 on MobileNetV2 and 2.3 on ViT-Tiny), and on this architecture its aggressiveness was not repaid. Across all settings, Opus ended within 0.3~pp of the 80-epoch recipe in 6 of 12 runs and within 0.5~pp in 7, against 9 and 11 for the rule, so its larger saving came with a less reliable accuracy. All methods gained on ViT-Tiny, with frontier gaps of +0.62~pp for Opus and Sonnet and +0.80~pp for the rule (4 wins in 4). These gains rest on frontiers that are either still rising or noisy. On CIFAR-10 the 60-epoch ViT-Tiny recipe is 0.91~pp below the 80-epoch one, against 0.23 and 0.16~pp for ResNet-18 and MobileNetV2. The CIFAR-100 ViT-Tiny frontier is nearly flat between 20 and 30 epochs (53.23\% and 53.34\%), and its two 80-epoch runs have a standard deviation of 0.81~pp. The rule's gain there (+1.03~pp at 42 data epochs) came from its early finish at epoch 40, not from virtual epochs.

\begin{table*}[t]
\centering
\caption{Accuracy (\%), data epochs $D$ and frontier gap $\Delta_F$ (pp) per setting, two-seed means.}
\label{tab:settings}
\small
\setlength{\tabcolsep}{4pt}
\begin{tabular}{@{}llcccccccccc@{}}
\toprule
 & & Recipe & \multicolumn{3}{c}{Rule} & \multicolumn{3}{c}{Sonnet} & \multicolumn{3}{c}{Opus} \\
\cmidrule(lr){3-3}\cmidrule(lr){4-6}\cmidrule(lr){7-9}\cmidrule(lr){10-12}
Dataset & Architecture & Acc. & Acc. & $D$ & $\Delta_F$ & Acc. & $D$ & $\Delta_F$ & Acc. & $D$ & $\Delta_F$ \\
\midrule
CIFAR-10 & ResNet-18 & 94.47 & 94.38 & 66.0 & $+0.06$ & 94.39 & 61.5 & $+0.15$ & 93.59 & 47.0 & $-0.31$ \\
 & MobileNetV2 & 92.73 & 92.56 & 66.0 & $-0.07$ & 92.48 & 64.5 & $-0.18$ & 92.48 & 51.0 & $-0.02$ \\
 & ViT-Tiny & 83.15 & 82.89 & 62.0 & $+0.57$ & 83.31 & 55.5 & $+1.20$ & 82.75 & 62.0 & $+0.43$ \\
\addlinespace
CIFAR-100 & ResNet-18 & 76.25 & 75.63 & 58.0 & $+0.13$ & 75.74 & 67.0 & $-0.11$ & 75.23 & 51.0 & $-0.09$ \\
 & MobileNetV2 & 72.20 & 72.16 & 66.0 & $+0.03$ & 71.97 & 72.0 & $-0.18$ & 72.27 & 52.5 & $+0.25$ \\
 & ViT-Tiny & 55.17 & 55.20 & 42.0 & $+1.03$ & 55.05 & 62.0 & $+0.03$ & 55.75 & 59.0 & $+0.81$ \\
\midrule
\multicolumn{2}{@{}l}{Mean of 12 runs} & 78.99 & 78.80 & 60.0 & $+0.29$ & 78.82 & 63.8 & $+0.15$ & 78.68 & 53.8 & $+0.18$ \\
\bottomrule
\end{tabular}
\end{table*}

Paired with each other, Opus saved 12.5~pp more of the data epochs than Sonnet (95\% CI 2.8 to 22.2; 11 wins in 12; Wilcoxon $p=0.011$) at the same frontier gap (+0.02~pp, $p=0.79$). Opus and the rule could not be told apart. Opus saved 7.8~pp more of the data epochs (95\% CI $-2.6$ to $+18.2$; $p=0.31$), and its frontier gap was 0.12~pp lower ($p=0.34$).

The a-priori rule shows what the same machinery achieves without the agents' choice of update. Its extrapolations were rejected in 70 of 92 attempts, in all six settings; the median rejected candidate lowered validation accuracy by 9.4~pp, and each rejection returned the slot to a data epoch. It accepted 22 virtual epochs in 12 runs, all of them in the last window (epochs 71 to 80). There the learning rate is close to zero, and an extrapolation of the last step barely moves the weights: the median relative change of the weights was $4\times10^{-4}$, against $9\times10^{-2}$ for the accepted virtual epochs of Opus. These are near-identity epochs in the sense of Section~\ref{sec:res-behaviour}. The a-priori rule saved 5.4\% of the data epochs and 0.9\% of the GPU time (Table~\ref{tab:pooled}), and more than half of that saving came from one CIFAR-100 ViT-Tiny run whose early finish fired at epoch 40. Paired with it, Opus saved 27.4~pp more of the data epochs (95\% CI 18.5 to 36.3; 11 wins in 12; Wilcoxon $p=0.001$) at the same frontier gap (+0.05~pp, $p=0.91$). Its accuracy against the 80-epoch recipe was 0.34~pp lower (95\% CI $-0.66$ to $-0.01$; $p=0.077$), which the frontier attributes to its 22 fewer data epochs. The a-priori rule stayed accurate because the acceptance rule caught its failures, not because it found a substitute for data epochs.

\subsection{What the agents did}
\label{sec:res-behaviour}

Opus chose averaging at its first analysis point. At epoch 10 it typically measured that consecutive epoch gradients were nearly uncorrelated, tried an extrapolation of the trajectory that lowered validation accuracy, and tried the mean of recent snapshots, which raised it. In the run of Appendix~\ref{app:program} the mean of the last five snapshots scored 88.66\% on the validation set, against 84.28\% for the current weights and 48.22\% for a one-epoch extrapolation. The notes Opus left for its next invocation called the resulting program an ``SWA-style average''. Opus tried averaging in the sandbox at the first analysis point in 11 of 12 runs, submitted an averaging program at that point in 10 and by the third analysis point in all 12, and every run accepted averaged virtual epochs. Of its 81 programs with V slots, 80 averaged snapshots (20 of them added a fraction of the recent weight drift to the mean) and one took a step along the last weight change. Eighty of the 81 selected between two or three candidates on the validation set, five weighted the snapshots unequally, and none used the gradients, the optimiser buffers or a per-layer rule. Of its 301 accepted virtual epochs, 284 were averages: of three snapshots in 131, five in 88, four in 35, two in 17, seven or eight in 11 and an unrecorded number in 2. The other 17 left the weights unchanged.

Every invocation of the main study used the same prompt, and each run started without notes from any other run. Averaging had, however, first appeared in a pilot that preceded the main study. The pilot used an earlier version of the protocol, in which virtual epochs could only open a window and the run length was fixed, on CIFAR-10 with ResNet-18 and seed~0 (the same split as the main runs). A fixed control that extrapolated the weight trajectory had all seven of its virtual attempts rejected and ran 80 data epochs. Opus averaged weight snapshots from its first analysis point, in all seven programs, and saved 15 data epochs at 94.31\%, against 94.50\% for the 80-epoch recipe; Sonnet adopted averaging at epoch 40 and saved four. We designed the rule, the window plan and the horizon after this pilot. The pilot's prompt may have contributed to the choice. We did not archive its text, but our design notes list snapshot averaging late in training among its guidance topics, and the reading guide of its summary table advised, for noise-dominated epoch gradients, to ``average them, move less, or ask for data''. For the main study we removed all such guidance (Section~\ref{sec:programs}), so our evidence that Opus chose averaging from its own measurements is that all 12 runs under the neutral prompt arrived at it independently.

Sonnet behaved differently. In 22 of its 83 windows, including 10 of 12 first windows, it planned data epochs only. Its 61 programs with V slots took an SGD step along recent epoch gradients (27), replayed the momentum buffer or the Adam moments (19) or extrapolated the weight trajectory (12). Only one run (CIFAR-10, ViT-Tiny, seed~0) averaged snapshots, from the fourth window on. Most programs scaled their step with the learning rate (55 of 61), and 23 restricted the update to the classifier head or the stem. The steps were small. The median accepted virtual epoch of Sonnet moved the weights by 0.5\% of the distance a data epoch covered at the same point of training, against 73\% for Opus. Of Sonnet's accepted virtual epochs, 79\% moved the weights by less than 5\% of that distance and another 7\% not at all (Table~\ref{tab:behaviour}). Sonnet's own notes say so. At epoch 40 of one run it wrote that its step ``is a near-no-op (update\_rel ${\sim}$5e-4) that reliably passes the acceptance band'' and planned four such epochs for the next window. Such an epoch amounts to skipping a data epoch while the learning-rate schedule moves on, and epochs of this kind account for 12.8 of the 16.2 data epochs Sonnet saved per run.

\begin{table*}[t]
\centering
\caption{Accepted virtual epochs and the source of the data epochs saved per run.}
\label{tab:behaviour}
\small
\setlength{\tabcolsep}{4pt}
\begin{tabular}{@{}lcccccccc@{}}
\toprule
 & & & \multicolumn{3}{c}{Size relative to a data epoch} & \multicolumn{3}{c}{Saved per run by} \\
\cmidrule(lr){4-6}\cmidrule(lr){7-9}
Method & Accepted V & Averages (\%) & Median & ${<}5\%$ (\%) & Median $\Delta A_{\mathcal{V}}$ (pp) & Substantial V & Near-identity V & Early end \\
\midrule
Rule & 150 & 49 & 0.135 & 49 & $+0.00$ & 6.4 & 6.1 & 7.5 \\
Sonnet & 178 & 3 & 0.005 & 87 & $-0.02$ & 2.0 & 12.8 & 1.4 \\
Opus & 301 & 94 & 0.727 & 6 & $+2.78$ & 23.5 & 1.6 & 1.2 \\
\bottomrule
\end{tabular}
\end{table*}

Table~\ref{tab:behaviour} measures the size of an accepted virtual epoch as the distance it moved the weights, divided by the distance a data epoch covered at the same clock epoch of the same setting. We took the latter from the per-group tables that the agents printed in their transcripts (745 measurements), and we call a virtual epoch near-identity when this ratio is below 5\%. Opus's virtual epochs moved the weights by 73\% of a data epoch in the median, raised validation accuracy by 2.78~pp and account for 23.5 of its 26.2 saved data epochs per run. The shares depend on the 5\% threshold, but the ordering of the three methods does not: with a threshold of 1\%, 6\% of Opus's virtual epochs are near-identity, against 67\% of Sonnet's and 45\% of the rule's.

\begin{figure*}[t]
  \centering
  \includegraphics[width=\textwidth]{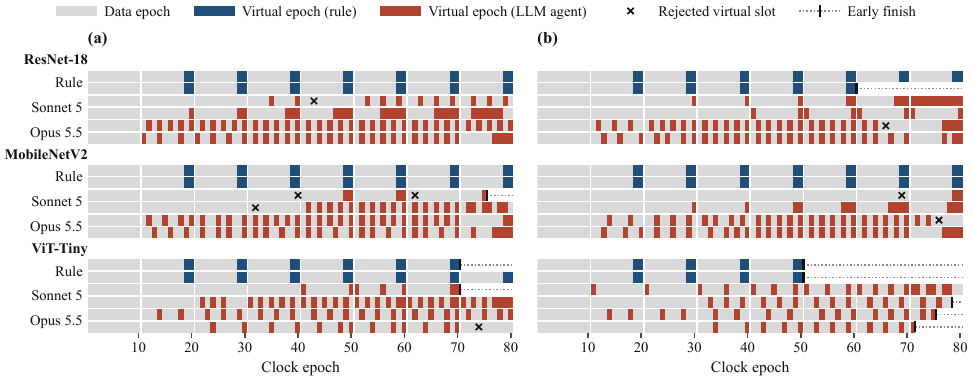}
  \caption{Epoch kinds in every run: (a) CIFAR-10, (b) CIFAR-100.}
  \label{fig:timeline}
\end{figure*}

The plans show how Opus adapted the timing (Fig.~\ref{fig:timeline}, one row per run). Up to epoch 70 it never planned two virtual epochs in a row, so every one of its V slots in windows 1 to 6 lay between data epochs. In windows 3 to 6, 5 to 7 of its 12 runs alternated D and V strictly, and Opus planned 4.1 virtual epochs per window, against 2.5 in the first window. In one run it wrote at epoch 20 that it was ``raising the V rate to alternating DVDV, using the same proven program''. In another run it noted that it avoided consecutive V slots because averaging over averages gives no progress. In the last window 7 of 12 plans contained consecutive V slots and all 12 ended on V; in 8 of the 12 runs the final weights were an accepted average, as in SWA. The number of Sonnet's virtual epochs grew from 0.2 per window in the first window to 4.4 in the last. The rule's fixed plan ends every window with two virtual epochs. No data epoch runs between them, so the second slot finds the same snapshots, and in 68 of its 75 windows it re-accepted the state of the first slot and changed nothing. Of the rule's four candidates, an extrapolation was kept in only 8 of 150 virtual epochs, which agrees with what the agents measured in the sandbox.

\begin{figure*}[t]
  \centering
  \includegraphics[width=\textwidth]{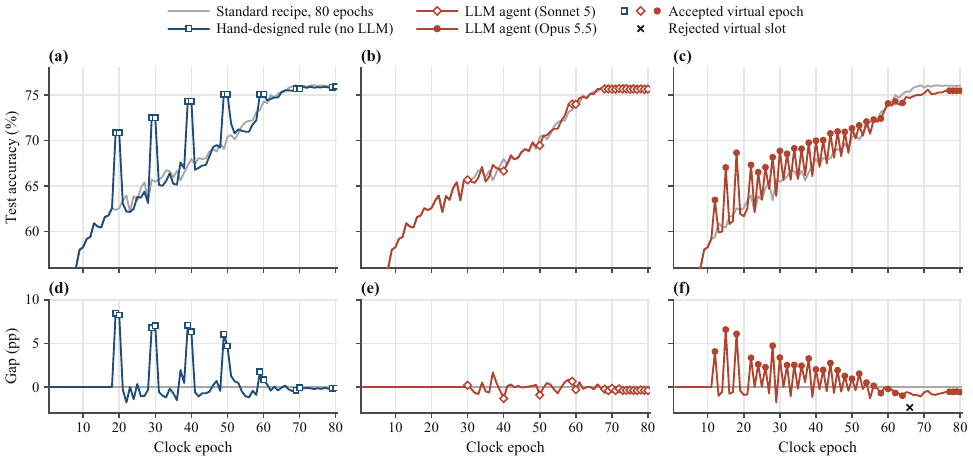}
  \caption{CIFAR-100, ResNet-18, seed~0: test accuracy (top) and gap to the 80-epoch recipe (bottom).}
  \label{fig:curves}
\end{figure*}

Fig.~\ref{fig:curves} follows one run of each method on CIFAR-100 with ResNet-18 (columns: rule, Sonnet, Opus). Early in training each accepted average raised test accuracy at once. The first virtual epoch of a rule window gained up to 9.4~pp (8.3~pp at epoch 19), and the averages of Opus gained 2.7 to 7.5~pp up to epoch 50 and less as the learning rate fell. The gain was gone after the next data epoch at the high learning rate, and accuracy then followed the curve of the 80-epoch recipe. Averaging therefore did not put training on a better trajectory. It replaced data epochs while the schedule advanced, with little lasting effect for the rule and a deficit of 0.6~pp for Opus at the end of this run. The virtual epochs that Sonnet placed at the end of this run, at low learning rates, changed almost nothing. Opus ended the run at 75.48\% with 51 data epochs, the rule at 75.94\% with 66 and Sonnet at 75.65\% with 62, against 76.08\% for the 80-epoch recipe of this seed.

Failed slots were rare. Of the V slots planned, 96\% of Opus's (301 of 312) and 92\% of Sonnet's (178 of 194) were accepted, and all 150 of the rule's. Most of the others never ran, because they followed a failure in the same window. Opus's programs declined two slots themselves because no candidate cleared their own margin, and Sonnet's declined four after a self-check against the acceptance rule. The acceptance rule itself rejected one candidate of each model. Opus shortened two runs, both on CIFAR-100 with ViT-Tiny at epoch 70: one ended at epoch 75, and the other after a single further epoch that returned what Opus called an ``SWA-like final model''. Sonnet shortened three runs, by 2 to 10 epochs. The rule's early-finish test fired seven times and shortened four runs, among them both CIFAR-100 ViT-Tiny runs, which ended at epoch 50 with 42 data epochs. The horizon accounts for 7.5 of the 20.0 data epochs the rule saved per run, against 1.2 of 26.2 for Opus.

\subsection{Compute, latency and API cost}
\label{sec:res-cost}

\begin{figure*}[t]
  \centering
  \includegraphics[width=\textwidth]{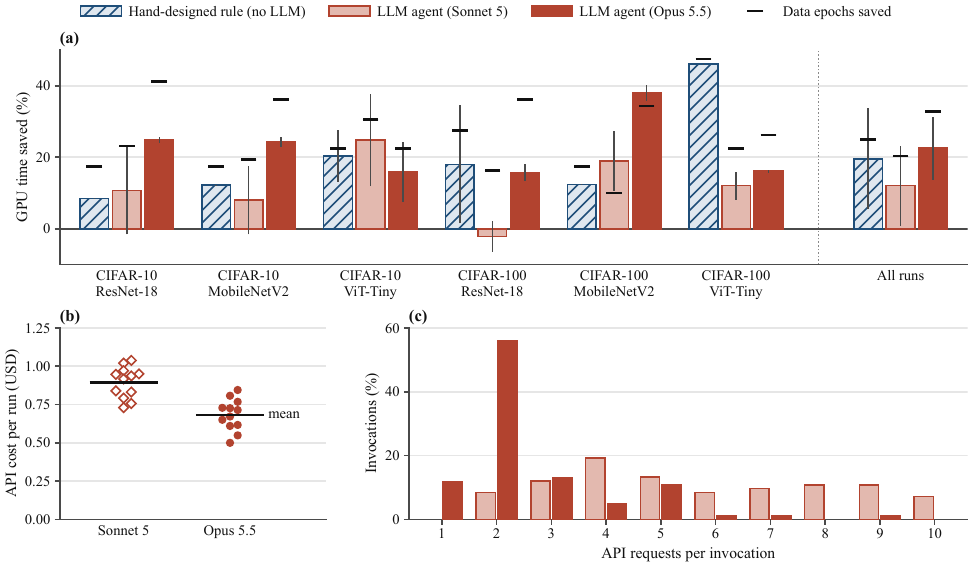}
  \caption{(a) GPU time saved; (b) API cost per run; (c) API requests per invocation.}
  \label{fig:cost}
\end{figure*}

Saved data epochs do not translate one for one into saved GPU time, because the method adds its own overhead (Fig.~\ref{fig:cost}a, where bars show one standard deviation over the runs of a group). With the calibrated times, Opus saved 22.5\% of the GPU time of the 80-epoch recipe (standard deviation 8.8), the rule 19.6\% and Sonnet 12.1\%, against 32.8\%, 25.0\% and 20.3\% of the data epochs. The differences of 10.3, 5.4 and 8.2~pp are the overhead of validation passes and recalibrations at every slot, the program calls and, for the agents, the recording of epoch gradients. One calibration entry is implausible: on CIFAR-100 with MobileNetV2, a data epoch that records gradients was timed faster than one that does not (4.95~s against 5.88~s), which flatters the agents in that setting. If every data epoch of the agents is priced at the larger of the two times, their savings fall to 20.8\% for Opus and 9.7\% for Sonnet, and their overhead rises to 12.0 and 10.6~pp.

The LLM's time is not GPU time, but in our implementation training waits for it. An Opus invocation took 29~s on average and a Sonnet invocation 86~s. Excluding the validation passes it requested, the agent added 3.0~min per run for Opus and 9.3~min for Sonnet, while an 80-epoch run of the recipe took 6.1~min of GPU time on average and Opus saved 1.4~min of it. Counting this latency, an Opus run took 39\% longer than the 80-epoch recipe and a Sonnet run 174\% longer. Over all runs, Opus spent 6.8~s of latency per saved data epoch and Sonnet 34~s, whereas a data epoch took 2.0 to 5.9~s in our settings, so at this scale the LLM costs more time than it saves.

An Opus run cost 0.68~USD in API fees (standard deviation 0.10) and a Sonnet run 0.89~USD (Fig.~\ref{fig:cost}b), although the list prices of Opus are twice those of Sonnet for input, output and cache-write tokens. Opus usually decided after two requests (56\% of invocations; mean 2.62, with 1.63 sandbox calls) and wrote 2{,}353 output tokens per invocation, thinking included. Sonnet needed 5.71 requests on average (six invocations reached the limit of ten), 4.73 sandbox calls and 7{,}164 output tokens (Fig.~\ref{fig:cost}c). Cache reads made up 77\% and 86\% of their input tokens. Per saved data epoch over all runs, Opus cost 0.026~USD and Sonnet 0.055~USD. The 24 agent runs of the main study cost 18.92~USD over 694 API calls, and with the pilots and the stopped runs of the earlier protocol the project spent 25.75~USD. In ten Sonnet invocations the sandbox audit flagged code that reached for functions outside the sandbox, mostly attempts to execute a draft program dynamically in order to test it; the sandbox blocked these attempts. No Opus invocation was flagged.

\section{Discussion and limitations}
\label{sec:discussion}

The results support a narrow claim. Given the weight and gradient history of a run and no hint about which update to try, Opus usually found during its first analysis that the mean of recent snapshots is a safe replacement for a data epoch that needs no training data. It then trained with a third fewer data epochs at a small loss of accuracy. That saving did not come with a gain over the recipe trained for the same number of data epochs. On ResNet-18 and MobileNetV2 all methods lay on the frontier, the gains on ViT-Tiny rest on frontiers that are still rising or noisy, and none of the pooled frontier gaps is distinguishable from zero in 12 runs. The time-to-target analysis agrees: against the best budget of the recipe, chosen in hindsight, none of the methods reached the target sooner by a margin we can resolve. In our experiments, a user who fixes a shorter budget in advance gets the same accuracy from the recipe alone, without an LLM. The meta-learner chooses its budget during the run, which a user of the recipe has to do beforehand, but we have not shown that this choice is better than a good default.

Two controls separate what the LLM contributes from what the machinery around it contributes. The a-priori rule has the same clock, plans, acceptance rule and budgets, with candidates from the closest prior method, and it saved 5.4\% of the data epochs. The rule differs from it only by the snapshot averages the agents had found, and it saved 25.0\%; Opus saved 32.8\%. The acceptance rule and the window plan therefore do not save data epochs by themselves. The choice of update does, and in this study that choice came from the LLM, which made it from the measurements of each run under a prompt that names no technique (Section~\ref{sec:programs}). Once the choice is known, a fixed rule implements it as well as Opus, without API cost or latency, and 12 pairs resolve no difference between the two. The contribution of the LLM is thus to find a working update inside a single run, where a designer would otherwise need pilot runs: the fixed extrapolation that prior work suggests failed in our pilot and was rejected in 70 of 92 attempts here. Beyond the choice of update, Opus set how often to apply it per setting (2.3 virtual epochs per window on ViT-Tiny against 4.4 on ResNet-18), declined slots when no candidate cleared its own margin and ended 8 of 12 runs on an average. Whether these adaptive choices beat a well-tuned fixed schedule, 12 runs cannot resolve, and on ResNet-18 the more aggressive schedule cost accuracy (Section~\ref{sec:res-frontier}).

This finding has a limit. Snapshot averaging was not new. SWA and LAWA~\cite{izmailov2018swa,kaddour2022lawa} were published years before the training cut-offs of both models~\cite{anthropic2026opus55docs,anthropic2026sonnet5docs}, Opus named its first program after SWA, and OPAL's agent also chose weight averaging in its own ResNet-18 experiment~\cite{bai2026reactive}. A designer who thought of averaging could have written the rule without an LLM, so the a-priori rule measures one plausible prior design, not every possible one. What the LLM supplied was the selection of a known technique, from among the candidates it tested in the sandbox, on the evidence of the current run. Knowledge of the technique alone did not produce this selection: Sonnet, whose training data cover the same methods, averaged in one run of twelve. Testing discovery beyond prior knowledge needs a setting in which the right update is not already known, such as label noise, class imbalance or other optimisers, and we leave it to future work.

The acceptance rule guards every step but not the result. It compares each candidate with the state the candidate replaces, so near-identity updates pass it, and so does a sequence of averages that each looks good but leaves the run behind the recipe. Opus's ResNet-18 runs accepted 124 virtual epochs, each of which passed the check, and still ended 0.94~pp below the full recipe, and Sonnet saved most of its data epochs with updates that it itself described as near no-ops. A rule that compared a candidate with a forecast of the data epoch it replaces, or a budget on the cumulative change of the validation loss, might prevent both, at the price of fewer accepted slots. The other choices of the protocol, namely the tolerance of 0.3~pp, the analysis period of ten epochs, the 80-epoch clock and the two levers, were fixed before the main runs and not tuned. The levers were designed after the pilot in which the agents had already found averaging.

The decisions of the LLM vary between runs of the same setting. Under the earlier protocol, the pilot and a later run with the same setting, seed and prompt used 65 and 73 data epochs with Opus. With Sonnet, one of these runs used 76 data epochs and averaged from epoch 40, and the other used 66 data epochs without any averaging program. Neither model accepts sampling parameters, so a seed cannot fix this variation, and two seeds per setting capture it only in part. Every accepted program is logged with its plan, so a sequence of decisions can be replayed without the LLM, but GPU training in reduced precision is not guaranteed to reproduce the same trajectory. Models may also be changed or retired by their provider; Sonnet~5 is already a legacy model~\cite{anthropic2026sonnet5docs}.

The evaluation is small. It covers three small architectures, two datasets of $32\times32$ images, an 80-epoch clock, two seeds per setting and two LLMs from one provider. Large-scale image training, language models and longer schedules may behave differently. We attribute the value of averaging here to small-batch SGD at a high learning rate, which moves the weights mostly through noise (consecutive epoch gradients had a cosine of about 0.03). The mean of recent snapshots then lies nearer the centre of the region that the iterates explore~\cite{izmailov2018swa}. How large that noise is in other regimes is open. The validation set is a class-balanced sample of the training distribution with 5{,}000 images, and it serves both for acceptance and for recalibrating batch normalisation. A smaller or shifted validation set would make acceptance noisier and could admit harmful states. For the same reason \method{} is not data-free in the sense of data-free distillation or quantisation, which use no real data at all. It removes gradient computations on the training set, not the need for held-out data, which the recipe also keeps for model selection.

At CIFAR scale the method costs more wall-clock time than it saves, because training waits for the LLM. The balance depends on how long a data epoch takes. Opus spent 6.8~s of latency per saved data epoch, so for models whose epochs take minutes the latency is minor, while the GPU overhead of validation and recalibration grows with the validation set but not with the training set. Training could also continue while the LLM analyses, with the previous program and plan; OPAL runs its agent in this way at 1.01 to 1.14 times the wall-clock time of training alone~\cite{bai2026reactive}.

\method{} executes code written by an LLM next to the training state. Our sandbox restricts names, imports and attributes, and it blocked every attempt in our runs to reach beyond them, but a restricted interpreter is not a security boundary, and a deployment should isolate the programs at the operating-system level. The LLM provider receives the brief and the printed output of the LLM's code. These contain summary statistics and can contain individual weight values, although full tensors never leave the machine. For proprietary models this disclosure may matter, and an open-weight model run on premises would avoid it, as would protocols that exchange queries instead of models~\cite{ruzbahani2026trustless}. More broadly, a channel that rewrites trained weights is an attack surface: adaptive attacks can defeat adversarially defended networks~\cite{MOHAMMADIRUZBAHANI2026104495}, and approximate unlearning shows how much direct weight edits can change what a model retains~\cite{ruzbahani2026tensorunlearn}.

\section{Conclusion}
\label{sec:conclusion}

\method{} lets an LLM decide, every ten epochs, which data epochs of a training run to replace with a training-data-free update and what that update is, with each decision written as a program and each replacement checked on a validation set. In 120 runs on CIFAR-10 and CIFAR-100, Claude Opus~5.5 used a third fewer data epochs than the 80-epoch recipe at 0.31~pp lower accuracy (up to 0.94~pp on ResNet-18, most of which the recipe also loses at that data budget). That places it on or slightly above the recipe's own trade-off between data and accuracy (+0.18~pp, not distinguishable from zero). Unprompted, it averaged recent weight snapshots in every run and interleaved the averages with data epochs. This choice produced the savings: the same machinery with update candidates from prior work saved 5.4\% of the data epochs, and with averaging added 25.0\%, as many as Opus within the resolution of 12 runs. The value of the LLM in this study lies in finding a usable update within a single run, at 0.68~USD per run, rather than in beating a fixed rule that already contains it. Open questions include settings in which the right update is not already known and larger models in which an epoch costs more than the LLM's latency. On the engineering side, invocations could run while training continues, acceptance rules could look beyond the next step, and open-weight models could keep the training state on premises.

\appendix

\section{An example decision of the LLM}
\label{app:program}

We describe the first decision that Opus submitted in the CIFAR-10 ResNet-18 run with seed~0, at epoch 10. Its first analysis step read the summary table, which showed a cosine of about 0.03 between consecutive epoch gradients. The second step scored three candidates on the validation set. The current weights reached 84.28\% validation accuracy at a loss of 0.4516. The mean of the last five snapshots reached 88.66\% at a loss of 0.327, a one-epoch extrapolation of the trajectory 48.22\% at a loss of 1.961, and a damped linear trend 78.54\% at a loss of 0.632. The summary of its reasoning that the API returned before this step reads: ``One option I'm weighing is averaging recent snapshots, which often helps validation at high rates, but averaging doesn't really make progress, so the final accuracy likely hinges on annealing.'' The phrase ``often helps'' shows that the choice drew on prior knowledge as well as on the measurements.

Opus then submitted a program of 20 lines with the rationale ``Averaging recent snapshots removes high-lr SGD noise: avg of W6..W10 gives 0.887/0.327 vs ref 0.843/0.452.'' At each virtual slot the program forms the means of the last three and of the last five snapshots and scores both on the validation set. It returns the one with the lower loss, and it declines the slot if neither is valid. It averages only the parameters and keeps the current normalisation buffers, which the training loop recomputes anyway. Opus submitted the same program with the plan DVDVDVDVDV at all seven analysis points of this run. All 35 virtual epochs were accepted, and the run ended at 93.66\% with 45 data epochs, 0.84~pp below the 80-epoch recipe of the same seed and 0.18~pp below its frontier. The programs and transcripts of all runs are released with the code.

\section*{Declaration of competing interest}

The authors declare that they have no known competing financial interests or personal relationships that could have appeared to influence the work reported in this paper.

\section*{Data availability}

CIFAR-10 and CIFAR-100 are publicly available~\cite{krizhevsky2009learning}. The code, configuration files, per-run logs, LLM transcripts and all programs produced in this study will be made publicly available upon publication.

\section*{AI-use disclosure}
During the preparation of this work the authors used AI in order to improve the readability of the code and to debug it, to improve the text, to correct grammatical errors, to trim the text, and to improve the visualisation of the figures. Separately, Claude Opus 5.5 and Claude Sonnet 5 are components of the method studied in this paper: they act as the meta-learner described in Sections~\ref{sec:method} and~\ref{sec:setup}, and all of their interactions during the experiments were logged.

\bibliography{refs}

\end{document}